\documentclass[letterpaper, 10 pt, conference]{ieeeconf}  

\IEEEoverridecommandlockouts                              
                                                          
\usepackage[utf8]{inputenc} 
\usepackage[T1]{fontenc}    
\usepackage{hyperref}       
\usepackage{url}            
\usepackage{booktabs}       
\usepackage{amsfonts}       
\usepackage{nicefrac}       
\usepackage{microtype}      
\usepackage{lipsum}
\usepackage{graphicx}
\usepackage{xcolor}

\title{Visual Goal-Directed  Meta-Learning with Contextual Planning Networks}

\author{
  Corban G. Rivera, David A Handelman\\
  Intelligent Systems Center\\
  Johns Hopkins Applied Physics Lab\\
  Laurel, MD 20723 \\
  \texttt{corban.rivera@jhuapl.edu} \\

}

\begin{document}
\maketitle

\begin{abstract}
The goal of meta-learning is to generalize to new tasks and goals as quickly as possible.  Ideally, we would like approaches that generalize to new goals and tasks on the first attempt.   Toward that end, we introduce contextual planning networks (CPN). Tasks are represented as goal images and used to condition the approach.  We evaluate CPN along with several other approaches adapted for zero-shot goal-directed meta-learning.  We evaluate these approaches across 24 distinct manipulation tasks using Metaworld benchmark tasks.  We found that CPN outperformed several approaches and baselines on one task and was competitive with existing approaches on others.  We demonstrate the approach on a physical platform on Jenga tasks using a Kinova Jaco robotic arm.

\end{abstract}


\section{INTRODUCTION}
When confronted with a new task, humans are able to generalize from previously learned skills and experience to perform new tasks on the first attempt. Equipping artificial intelligence (AI) agents with similar capabilities would increase their value as robotic teammates by enabling "improvisation" of problem solutions. If an AI could learn to transfer learned skills to new tasks in a way similar to humans, it would make the robotic teammate much more useful and trusted as a partner. It has been thoroughly studied that state-of-the-art AI struggles when presented with a limited number of  training examples of new tasks, let alone succeeding in a "zero-shot" task with no previous experience attempting it  \cite{ntp}.    One key challenge is learning abstractions and modularity that will allow novel tasks to be completed on the first attempt. 

There are several related problems that have received a lot of attention in literature including meta-learning and meta-imitation learning which we will describe briefly.

\textbf{Meta-learning}
Meta-learning, or "learning to learn," has a rich history in machine learning literature  \cite{learning_to_learn,godel,bengio_metalearning,l2l_by_gradient}.  
Given a family of tasks, training experience for each meta-training task, and performance measures, an algorithm is capable of learning to learn if its performance improves with both additional experience and tasks.  

Meta-learning generally refers to the problem of quickly adapting to new tasks given prior experience on dissimilar tasks.   These methods frequently include an outer loop which is used to select parameters for an inner loop which is then used to solve the task  \cite{learning_to_learn,bengio_metalearning}.  Some approaches rely on conditioning networks on a task embedding  \cite{pearl}.  Other approaches optimize directly for their ability to fine tune the model quickly to a library of known tasks  \cite{maml,anml}.

\begin{figure}[h!]
\centering
  \includegraphics[width=\linewidth]{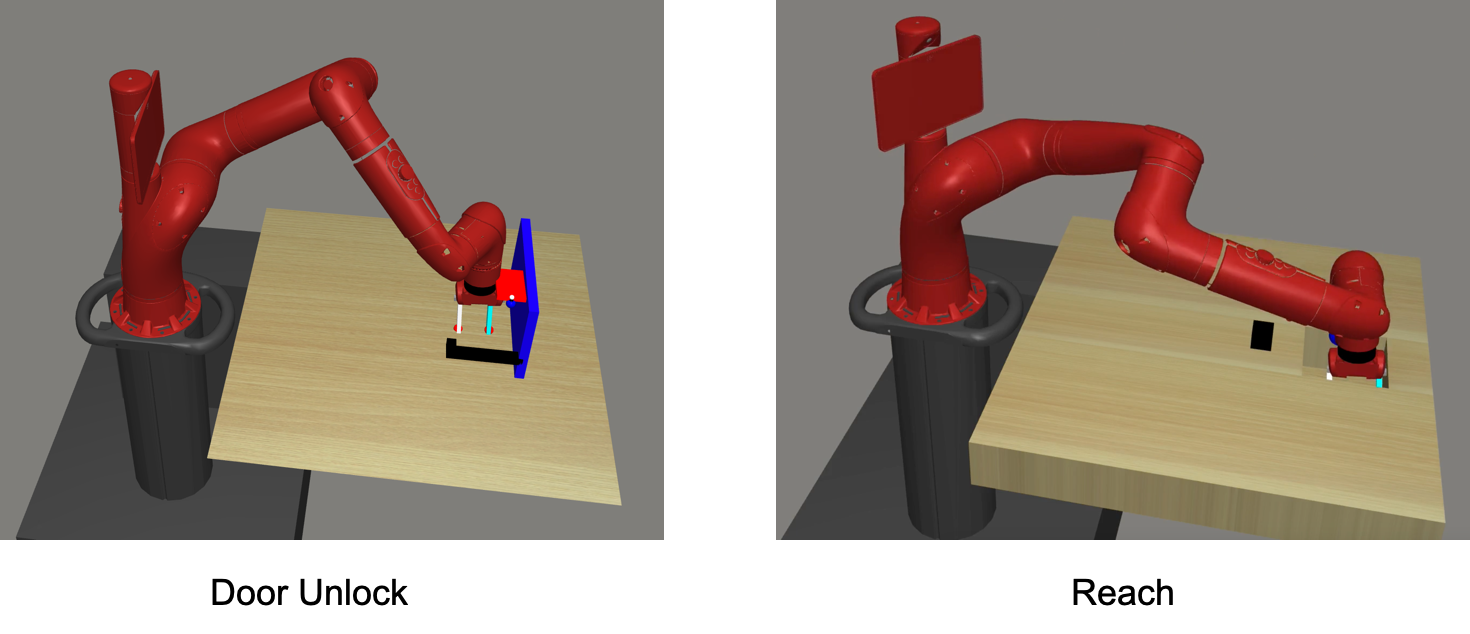}
  \caption{Metaworld is a manipulation benchmark that includes 50 distinct tasks for multi-task learning and meta-RL experiments. The tasks include distinct objects and affordances.  We adapted the metaworld benchmark tasks to create 24 zero-shot meta-learning from visual demonstration tasks for evaluation.  The use of visual observations and goals is slightly more challenging than the existing metaworld benchmark formulations which are based on coordinate-based observations and goals}
  \label{fig:metaworld}
\end{figure}

\textbf{Meta-imitation learning}
Imitation learning allows autonomous agents to learn a skill from demonstrations.  Neural network based policies can provide a basic level of generalization to new scenarios, however training these policies may require many demonstrations.  Research into few-shot learning and rapid adaptation has focused on decreasing the number of demonstrations or trials of the test-task needed to successfully perform the new task \cite{fewshot,maml}.  In the limit, zero demonstrations of the new task are provided.  This is a special case of  \emph{zero-shot learning} in policy domain, which has become an emerging sub-field within meta-learning  \cite{visual_planning}. \emph{Meta-imitation learning} refers to meta-learning approaches that use imitation learning during the meta-training phase.

\textbf{Contextual Planning Networks}
The problem that we are solving here is distinct from the standard formulations of meta-learning and meta-imitation learning in that we aim to complete a meta-test task on the first attempt assuming we are given demonstrations of the meta-training tasks and a single image of the completion of the meta-test task.
To solve this problem, We introduce an approach called contextual planning networks (CPN) that learns a combined representation of the policy and dynamics using an objective that directly optimizes for the ability to plan towards a goal state represented as an image. Our approach draws inspiration from several previous approaches  \cite{upn,pearl,anml} and builds upon them. The approach combines planning by backpropagation, with task embedding, and neuromodulation.

\textbf{Task Objectives given as an Image}
We focus on approaches that represent their goal as an image.  The idea of using an image as a goal has been studied previously  \cite{goalimage,latentimages}. An image provides a flexible format that can be easily accommodate diverse tasks. The challenge comes from finding representations of the goal image that allow the policy to accomplish the test task. 
Using an image as a task goal has unique challenges as many pixel level details of the image may be irrelevant or misleading for the task.  Instead, many approaches have explored latent representations of the images that are intended to capture the salient aspects of the task from the image  \cite{latentimages,upn}.  To discover these latent representations, prior work has mostly focused on unsupervised objectives or objectives that are disconnected from learning the policy  \cite{latentimages,finn2016deep}.

\textbf{Planning by Backpropagation}

The strength of planning for the purpose of meta-learning is the ability to transfer an existing policy towards a new task and potentially complete a previously unseen goal.  

Planning through latent space with gradient descent is a core concept shared by several previous works, in which the same embedding is used for both the goal and the current state. The basic idea is that the distance between the latent representation of the current state and the latent representation of the goal could be exploited for planning purposes  \cite{visual_planning}.  The idea can be extended by using a dynamics model to predict the latent state resulting from an action.  Prior work   \cite{upn,oh2017value,silver2017predictron,muzero,pathak2018zero} has demonstrated the value of using the dynamics model to unroll the policy over a planning horizon with the goal of minimizing the distance between the latent representations of the final predicted state and the goal.

The primary contribution of this paper is a visual goal-directed meta-learning approach that allows existing policies to be repurposed for new tasks with a chance of succeeding on the first attempt. Our experiments show that the approach improves the task success rate on the first attempt for one task in the metaworld \cite{metaworld} task library  and is competitive with recent approaches and baseline algorithms for many others.  In a later experiment, we show that the approach works on a physical robot using a Kinova Jaco robotic arm and tasks based on the game Jenga.

\section{RELATED WORK}
 Describing tasks by images has been studied by several authors \cite{finn2016deep,latentimages}.  These approaches do not focus on zero-shot goal-directed task transfer.  Other work used an imitation learning objective to optimization a representation for a goal image  \cite{sermanet2016unsupervised,sermanet2017time}. In contrast, our work focuses on both imitation learning and planning.  Single-shot imitation learning via images has also been explored \cite{nair2017combining,zhou2019watch,ntp,taco}.  Single-shot visual imitation learning approaches are related in the sense that the final image of the single demonstration could be thought of as the goal image.

Meta-learning in the context of reinforcement learning is referred to as Meta-RL \cite{metarl,rl2,maml,anml,pearl}. These methods are related in that they evaluate the approaches on their ability to learn new tasks with as few steps from the new task as possible.  They are different in that they require environments with extrinsic rewards.  In contrast, the work here is based on meta-imitation learning.   Although, inverse reinforcement learning  \cite{irl,irl2} (IRL) could be used to approximate an objective function which would allow Meta-RL approaches to be applied for learning from demonstrations. Instead of taking the IRL coupled with Meta-RL approach, In our experiments, we  adapted  concepts from Meta-RL approaches into the visual learning from demonstration context explicitly for the purpose of comparison with a baseline approach that conditions the policy on both a state and task-embedding \cite{pearl}.   

The earliest work on planning by gradient descent was introduced decades ago  \cite{kelley1960gradient} and was based on known model dynamics. Later work introduced planning with learned dynamics models \cite{schmidhuber1990line}.  Model-based planning was also explored for environments with discrete actions \cite{henaff2017model}.  This approach relies on unsupervised pretraining.  Others have explored using planning through latent space with the goal of predicting a value function \cite{oh2017value,silver2017predictron,muzero}. These approaches focus on environments with extrinsic rewards. Other work has looked at approaches for goal-directed imitation learning \cite{pathak2018zero} conditioned on a sequence of images of the test task.

\section{GOAL-DIRECTED ZERO-SHOT META-LEARNING}

We first introduce some meta-learning preliminaries, then we present the problem statement before describing the approach and implementation.

\subsection{PRELIMINARIES}
Meta-learning, or "learning to learn," aims to complete new tasks with very little (if any) training data. To achieve this, a meta-learning algorithm is first pretrained on a repertoire of tasks ${\mathcal{T}_i}$ in a meta-train phase.  The algorithm is then evaluated on a disjoint task $\mathcal{T}_j$.  As part of meta-training, the meta-learning algorithm can learn generalizable structure between tasks that allow it to successfully complete the meta-test task.

A task $\mathcal{T}_i$ is a finite-horizon Markov decision process (MDP), $\lbrace{S,A,r_i,P_i,H\rbrace}$ with state space $S$ given as images, action space $A$, reward function $r_i : S \times A \rightarrow \mathcal{R} $, dynamics $P_i(s_{t+1}|s_t,a_t)$, and horizon $H$. It is assumed that the environment dynamics and goals may vary across tasks.

We explore a subset of reward specification $\mathcal{R}$ corresponding to the mean squared error  $||f_\phi(s_t)-f_\phi(s_g)||_2 $ between state $s_t \in S $ at time $t$, a goal $s_g \in S$, and image embedding function $f_\phi$ with parameters $\phi$.

\subsection{PROBLEM STATEMENT}
The goal is to meta-train an agent such that it is more successful at performing a new test task $\mathcal{T}_j$ on the first attempt given a goal $g_j$ for test task $j$ presented as an image.  

\textbf{Meta-train:} The agent observes and learns from $m = 1,\ldots,M$ task demonstrations.

\textbf{Meta-test:} The pretrained agent is given a new goal $g_j$ that corresponds to the goal for the test task $\mathcal{T}_j$.

\subsection{UNIVERSAL PLANNING NETWORKS}
\begin{figure}[h!]
\centering
  \includegraphics[width=\linewidth]{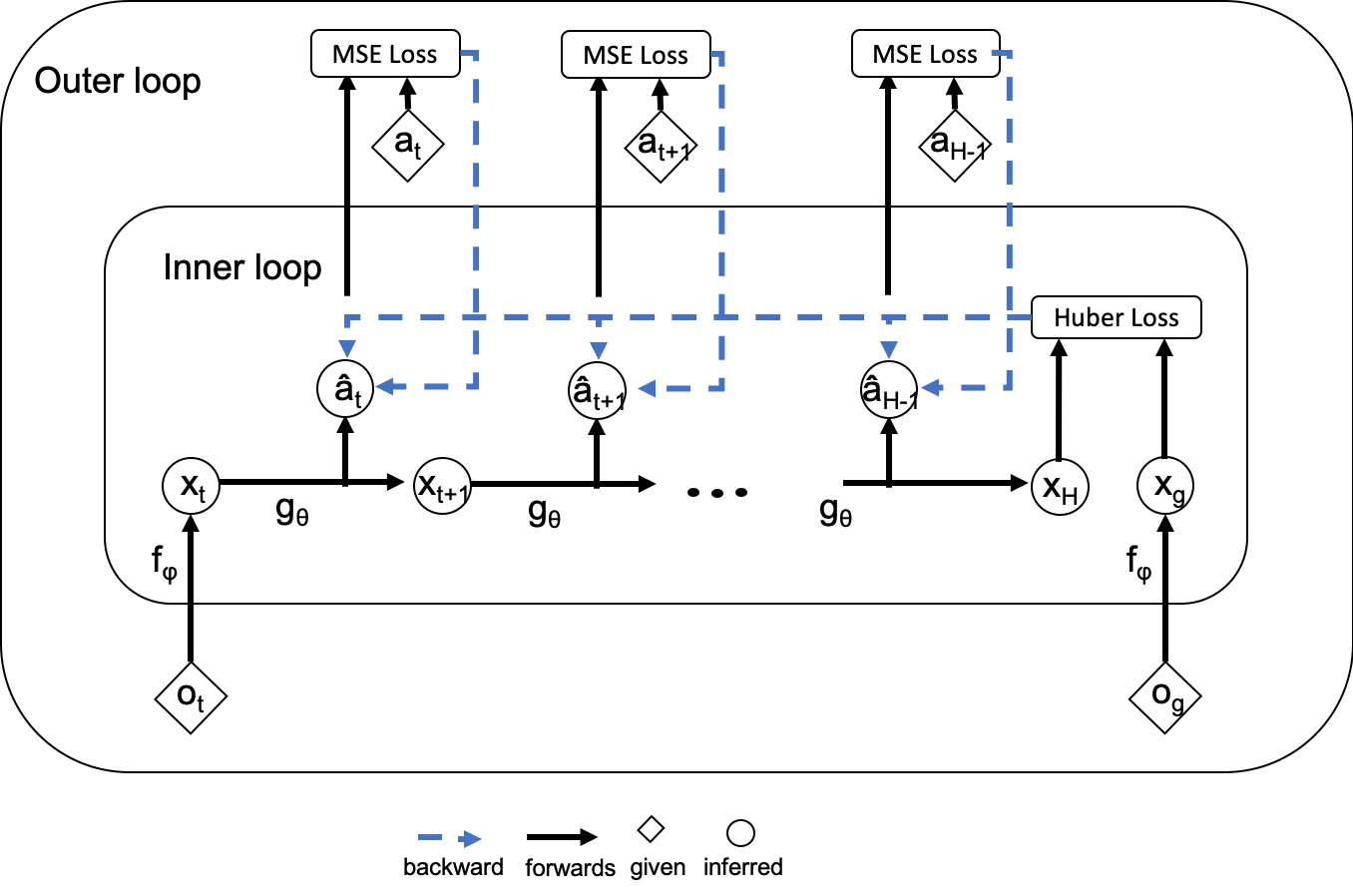}
  \caption{Universal planning networks architecture.  The architecture can be described as an inner loop and an outer loop.  The objective of the inner loop is to minimize the distance in latent space between the final rolled out state prediction and the latent representation of the goal. For learning from demonstrations, the outer loop is the behavior cloning objective which minimizes the distance between the predicted and observed actions.}
  \label{fig:upnarch}
\end{figure}

The goal of this meta-learning approach is to directly optimize for plannable representations \cite{upn}.  The approach is inspired by other well known approaches like MAML \cite{maml} that optimize a model for an ability to quickly fine tune for a new task. The UPN approach is illustrated in Figure \ref{fig:upnarch}. The model produces an action plan $\hat{a}_{t:t+H}$ conditioned on initial and goal observations $o_t$ and $o_g$, where $\hat{a}_t$ denotes the predicted action at time $t$ over horizon $H$.  Both the encoder $f_\phi$ and the combined policy and dynamics model $g_\theta$ are approximated with neural networks and are fully differentiable.  The inner-loop unrolls predictions of intermediate states $x_t:x_{t+H}$ and actions $\hat{a}_{t:t+H}$ over the horizon $H$.  The objective of the inner loop is to minimize the distance between the latent representation of the goal $x_g$ and the final predicted state $x_{t+H}$  using the Huber loss.  Updates to the model using this loss result in updated actions $\hat{a}_{t:t+H}$.  In the learning from demonstrations scenario, the outer loop uses a behavior cloning objective that minimizes the distance between the predicted action $\hat{a}_t$ for time $t$ and the given action $a_t$ over all timepoints using a mean squared error loss.  The planning horizon $H$ and the number of inner-loop updates $U$ are hyperparameters for the approach.  In our experiments, hyperparameters are set such that the predicted state embedding at the end of the horizon is as close as possible to the embedding of the goal.

\begin{figure}[h!]
\centering
  \includegraphics[width=\linewidth]{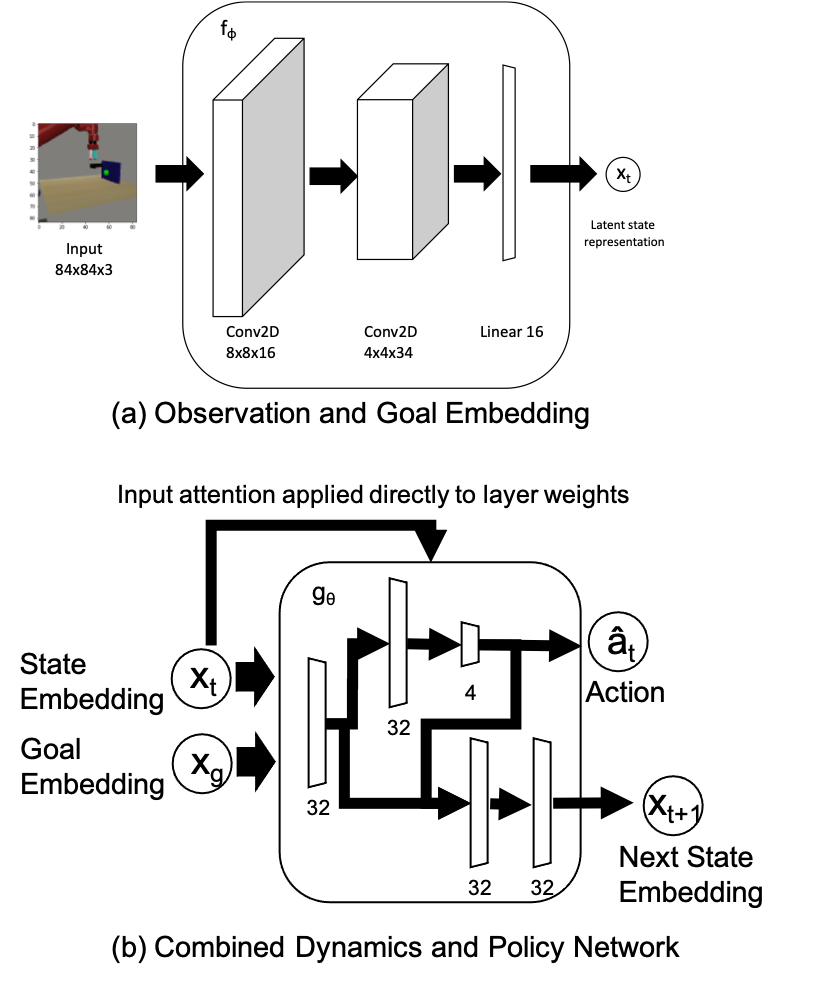}
  \caption{(a) Network architecture of the observation and goal embedding, (b) Structure of the combined policy and dynamics model.  A key structural difference between this model and an MLP architecture is the introduction of attention conditioned on the input that attenuates the weights of the model directly.  }
  \label{fig:encoder}
\end{figure}

\subsection{NEUROMODULATED META-LEARNING}
Neuromodulation in deep networks is inspired by the plasticity response observed for human neurons.  In our approach, this is achieved by learning and applying attenuation to network weights directly conditioned on the input \cite{anml}.  This is in contrast to commonly used notions of attention that learn and apply attenuation directly to the input conditioned on the input \cite{attention}. Figure \ref{fig:encoder} illustrates the network architecture of the of the combined dynamics and policy network.   

Given fully connected layers of the form $y=ax+b$, we introduce weight attenuation functions $u_\beta(x)$ and $v_\gamma(x)$ conditioned on the input $x$ .  The function $u_\beta(x)$ is parameterized by $\beta$ and yields a tensor that matches the shape of $a$ with values in the range $[0-1]$.  The bias attenuation function $v_\gamma(x)$ is parameterized by $\gamma$ and yields a matrix that matches the of size $b$ in the range $[0-1]$.  The neuromodulated fully connected layer computes $y = (u_\beta(x)\odot a)x+(v_\gamma(x) \odot b) $ where $\odot$ denotes element-wise multiplication.

In our experiments, Both $u_\beta(x)$ and $v_\gamma(x)$ are approximated using a 2-layer fully-connected neural network with intermediate rectified linear activation and terminal sigmoid activation.

\subsection{TASK-EMBEDDING META-LEARNING}
Task embedding in reactive networks has been explored for meta-learning \cite{pearl,pathak2018zero}. A recent approach called PEARL \cite{pearl} extends the typical policy formation $p(a|s)$ by additionally conditioning on a latent representation of the task $p(a|s,x_g)$.  An additional KL-divergence objective is used to organized the task embedding into a desired distribution.  This approach was demonstrated in the context of reinforcement learning.  We adapted this approach to learn from demonstration by adopting a behavior-cloning objective. In experiments, we refer to this baseline approach as (TE-BC) for task embedding behavior cloning.  

\subsection{CONTEXT-AWARE PLANNING NETWORKS}
In this paper, we introduce context-aware planning networks with extends the ideas from universal planning networks illustrated in Figure \ref{fig:upnarch} with neuromodulation illustrated in Figure \ref{fig:encoder}(b) and visual task-embedding illustrated in Figure \ref{fig:encoder}(a). Task-embedding is integrated by modifying the combined policy and dynamics model by additionally conditioning on $x_g$ as shown in Figure \ref{fig:encoder}(b).  Neuromodulation is integrated into the full model by replacing fully connected layers used to approximate $g_\theta(x_t,x_g)$ with neuromodulated fully-connected layers as described earlier.  The structure of combined dynamics and policy network has a shared trunk that passes the image embedding to a linear layer .  After the shared trunk, the network splits into dynamics and policy branches.   The policy branch is composed of linear layers and the number of continuous actions.  The dynamics branch is composed of two linear layers and is also conditioned on the action prediction.  In experiments with metaworld, the number of continuous actions is 4 and the internal linear layers are 32 units wide.

\section{RESULTS}

We designed our experiments to answer the following questions:  (i) Conditioned on a goal image, how effective is the CPN approach at completing meta-test tasks on the first attempt compared to UPN and imitation learning and task embedding approaches?  (ii) Can the approach work on a physical robot?

\subsection{Methods for comparison}
We compared to two reactive imitation learning approaches and two planning approaches along with a random control.  We included a behavior cloning based approach (BC), which uses a sequence of convolutional layers to encode visual observations and output action.  We compare to a Task Embedding Behavior Cloning (TE-BC) approach that extends the behavior cloning model by additionally conditioning on an embedding of the goal image.  In this approach, both the observation and the goal image are processed by the same learned convolutional encoding network.

Additionally, we compare to two approaches that plan over a finite horizon by gradient decent.  The first approach is based on the Universal Planning Network  with a horizon of 5 (UPN) \cite{upn}.  We also compare to an approach that extends UPN with neuromodulation for the integrated policy and dynamics model along with task embedding (TE-CPN).  A single backpropagation update step was used by each planning algorithm to plan towards the goal.  For these experiments the planning horizon was fixed at 5.  Both planning based approaches were evaluated based on the strategy of model predictive control described by Srinivas et al. \cite{upn}, which amounts to replanning actions over a fixed horizon at each time step but only taking the first action in the plan.

\subsection{Metaworld tasks}
We structured our experiements around 24 distinct visiomotor tasks derived from the Metaworld benchmark \cite{metaworld}.  Two example tasks are shown in Figure \ref{fig:metaworld}.  The tasks contain different objects with different affordances.  Additionally, each task contains selectable sub-task variation.  The original metaworld benchmarks were designed for multitask and meta-RL with vector representations of state and goals.  We adapted the metaworld benchmark tasks to make them suitable for approaches that learn from demonstration with visual observations and goals.  The introduction of visual observations add additional complexity that are not part of the existing Metaworld experiments.

\subsection{Experiment design}
We adopted a cross-validation structure for the experiment. Specifically, to evaluate the first attempt success rate on a meta-test task, the meta-training set consisted of all other tasks.  

\textbf{Training data}
For each of the 24 tasks, we generated 100 successful demonstrations using heuristic control. For each trial, the position of key objects was randomized within preconfigured bounds.  Trajectories in the dataset consisted of 84x84x3 images of each state including the final state which was treated as the goal image.  Each approach was trained for 50 epochs.  

\textbf{Evaluation}
For the task being evaluated, we measured average task success rate over 20 trials.  We controled the random seed to ensure that each approach was evaluated based on the same distribution of task variants.

\begin{figure}[h!]
\centering
  \includegraphics[width=\linewidth]{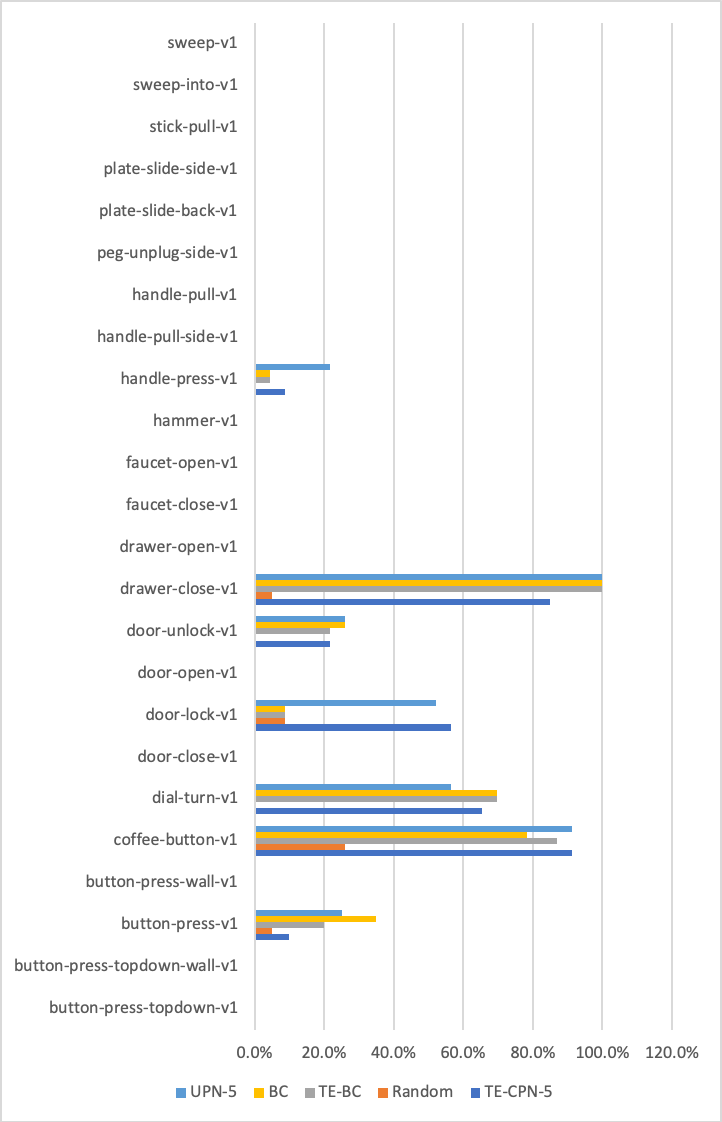}
  \caption{First attempt success rate for several meta-learning and baseline approaches based on reactive policies and planning.  We organized an experiment around the metaworld series of manipulation tasks.  For each evaluation task, the approaches were trained using 100 successful demonstrations from all of the other tasks.  Each approach was trained for the same number of epochs.  Average \% task success is on the x-axis. Evaluated tasks are shown on the y-axis.  We compare reactive imitation learning based on behavior cloning (BC), an approach that extends BC by conditioning the policy on both the observation and goal embedding (TE-BC), universal planning network (UPN), and contextual planning networks (TE-CPN) that extends UPN with neuromodulation and task embedding.  A random policy is included to illustrate the likelihood of task completion by chance. }
  \label{fig:first_attempt}
\end{figure} 

The results of the experiments are summarized in Figure \ref{fig:first_attempt}.  We found that 7 of the 24 evaluation tasks had at least one approach that was able to complete the task on the first attempt. For the door-lock task, CPN and UPN achieved over $50\%$ success on the first attempt, while the other reactive approaches performed close to chance.  For the drawer-close, dial-turn, coffee-button tasks all of the approaches performed well above chance. 

\subsection{JENGA domain on a physical robot}

We wanted to test the ability of CPN to extrapolate to novel goal targets on physical hardware.   Kinova Jaco arms and an oversized Jenga tower were used for the experiment.  The task was to demonstrate the ability to poke target blocks in the tower in positions that were were offset along an axis not seen during training.  

\begin{figure}[h!]
\centering
  \includegraphics[width=\linewidth]{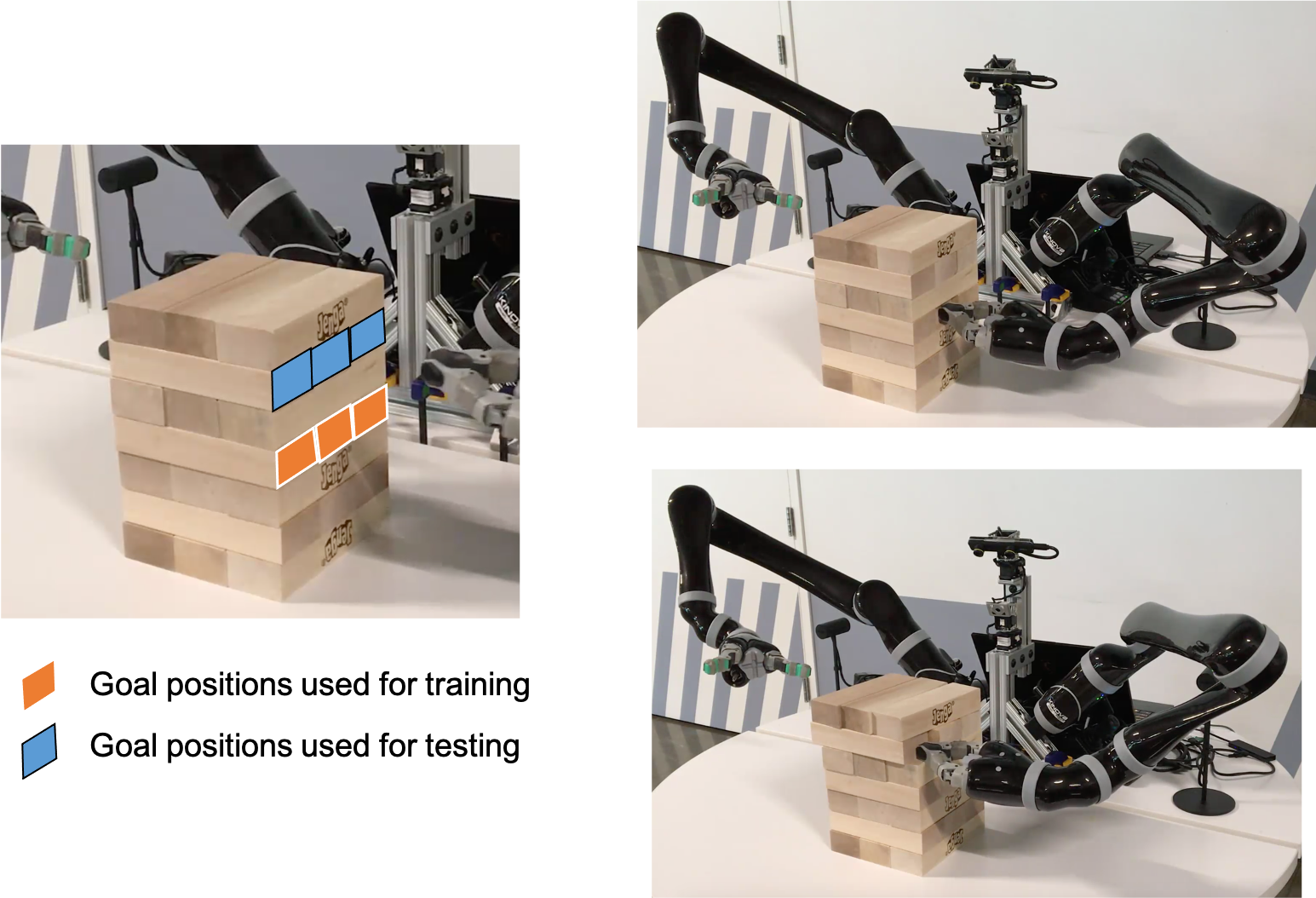}
  \caption{Physical demonstration of planning towards novel goals in Jenga.  CPN was trained using trajectories of Jaco arm reaching towards
  blocks on the lower level.  CPN was used to create a trajectory for a target block that were offset along an axis that was not varied during training.}
  \label{fig:jenga}
\end{figure}

Heuristic behaviors were used to create a dataset of 21 trajectories of the JACO arm poking a block on the lower row from different starting positions.  Figure \ref{fig:jenga} illustrates the experimental setup.  In this experiment, CPN was reconfigured to be conditioned on a block position goal represented as a three dimensional vector.  The horizon length and number of planning updates were selected to minimize the deviation between the predicted endpoint of the trajectory and the target block position.  In Figure \ref{fig:jenga} and the supplementary video, we show that the approach was able to plan towards multiple targets that offset along an axis not seen during training.  

\section{DISCUSSION}

Of the tasks that were solvable on the first attempt, we were surprised to find that there was no clear superior approach among the approaches that we tested.  The best approach frequently varied by task.  Occasionally, basic reactive approaches fared as well as more sophisticated planning-based approaches.  We found that the CPN approach outperformed the other approaches in the door-lock task, and we found that the CPN was competitive with the other approaches for the other tasks in the metaworld experiment.

In the Jenga experiment, we found that CPN was successful at planning towards goals that were not seen during training.  This experiment highlights the ability of the CPN approach to extrapolate to new goal targets not seen during training.  The experiment represents extrapolation because the goal targets were offset along an axis that was not in the training data.  While we fully appreciate that the task could be accomplished via inverse kinematics, we hope the experiment illustrates our contribution to the more fundamental challenge of zero-shot goal-conditioned task planning.

\section{ACKNOWLEDGEMENTS}
We would like to thank Chace Ashcraft, I-Jeng Wang, and Marie Chau for technical and manuscript feedback.

\bibliographystyle{IEEEtran}
\bibliography{IEEEabrv,references}

\begin{thebibliography}{10}
\providecommand{\url}[1]{#1}
\csname url@rmstyle\endcsname
\providecommand{\newblock}{\relax}
\providecommand{\bibinfo}[2]{#2}
\providecommand\BIBentrySTDinterwordspacing{\spaceskip=0pt\relax}
\providecommand\BIBentryALTinterwordstretchfactor{4}
\providecommand\BIBentryALTinterwordspacing{\spaceskip=\fontdimen2\font plus
\BIBentryALTinterwordstretchfactor\fontdimen3\font minus
  \fontdimen4\font\relax}
\providecommand\BIBforeignlanguage[2]{{%
\expandafter\ifx\csname l@#1\endcsname\relax
\typeout{** WARNING: IEEEtran.bst: No hyphenation pattern has been}%
\typeout{** loaded for the language `#1'. Using the pattern for}%
\typeout{** the default language instead.}%
\else
\language=\csname l@#1\endcsname
\fi
#2}}

\bibitem{ntp}
D.~Xu, S.~Nair, Y.~Zhu, J.~Gao, A.~Garg, L.~Fei-Fei, and S.~Savarese, ``Neural
  task programming: Learning to generalize across hierarchical tasks,'' in
  \emph{2018 IEEE International Conference on Robotics and Automation
  (ICRA)}.\hskip 1em plus 0.5em minus 0.4em\relax IEEE, 2018, pp. 1--8.

\bibitem{learning_to_learn}
\BIBentryALTinterwordspacing
S.~Thrun and L.~Pratt, \emph{Learning to Learn: Introduction and
  Overview}.\hskip 1em plus 0.5em minus 0.4em\relax Boston, MA: Springer US,
  1998, pp. 3--17. [Online]. Available:
  \url{https://doi.org/10.1007/978-1-4615-5529-2_1}
\BIBentrySTDinterwordspacing

\bibitem{godel}
J.~Schmidhuber, ``G{\"o}del machines: Fully self-referential optimal universal
  self-improvers,'' in \emph{Artificial general intelligence}.\hskip 1em plus
  0.5em minus 0.4em\relax Springer, 2007, pp. 199--226.

\bibitem{bengio_metalearning}
S.~Bengio, Y.~Bengio, J.~Cloutier, and J.~Gecsei, ``On the optimization of a
  synaptic learning rule,'' in \emph{Preprints Conf. Optimality in Artificial
  and Biological Neural Networks}, vol.~2.\hskip 1em plus 0.5em minus
  0.4em\relax Univ. of Texas, 1992.

\bibitem{l2l_by_gradient}
S.~Hochreiter, A.~S. Younger, and P.~R. Conwell, ``Learning to learn using
  gradient descent,'' in \emph{International Conference on Artificial Neural
  Networks}.\hskip 1em plus 0.5em minus 0.4em\relax Springer, 2001, pp. 87--94.

\bibitem{pearl}
K.~Rakelly, A.~Zhou, C.~Finn, S.~Levine, and D.~Quillen, ``Efficient off-policy
  meta-reinforcement learning via probabilistic context variables,'' in
  \emph{International conference on machine learning}, 2019, pp. 5331--5340.

\bibitem{maml}
C.~Finn, P.~Abbeel, and S.~Levine, ``Model-agnostic meta-learning for fast
  adaptation of deep networks,'' \emph{arXiv preprint arXiv:1703.03400}, 2017.

\bibitem{anml}
S.~Beaulieu, L.~Frati, T.~Miconi, J.~Lehman, K.~O. Stanley, J.~Clune, and
  N.~Cheney, ``Learning to continually learn,'' \emph{arXiv preprint
  arXiv:2002.09571}, 2020.

\bibitem{fewshot}
Y.~Wang, Q.~Yao, J.~T. Kwok, and L.~M. Ni, ``Generalizing from a few examples:
  A survey on few-shot learning,'' \emph{ACM Computing Surveys (CSUR)},
  vol.~53, no.~3, pp. 1--34, 2020.

\bibitem{visual_planning}
K.~Liu, T.~Kurutach, C.~Tung, P.~Abbeel, and A.~Tamar, ``Hallucinative
  topological memory for zero-shot visual planning,'' \emph{arXiv preprint
  arXiv:2002.12336}, 2020.

\bibitem{upn}
A.~Srinivas, A.~Jabri, P.~Abbeel, S.~Levine, and C.~Finn, ``Universal planning
  networks,'' \emph{arXiv preprint arXiv:1804.00645}, 2018.

\bibitem{goalimage}
K.~Deguchi and I.~Takahashi, ``Image-based simultaneous control of robot and
  target object motions by direct-image-interpretation method,'' in
  \emph{Proceedings 1999 IEEE/RSJ International Conference on Intelligent
  Robots and Systems. Human and Environment Friendly Robots with High
  Intelligence and Emotional Quotients (Cat. No. 99CH36289)}, vol.~1.\hskip 1em
  plus 0.5em minus 0.4em\relax IEEE, 1999, pp. 375--380.

\bibitem{latentimages}
M.~Watter, J.~Springenberg, J.~Boedecker, and M.~Riedmiller, ``Embed to
  control: A locally linear latent dynamics model for control from raw
  images,'' in \emph{Advances in neural information processing systems}, 2015,
  pp. 2746--2754.

\bibitem{finn2016deep}
C.~Finn, X.~Y. Tan, Y.~Duan, T.~Darrell, S.~Levine, and P.~Abbeel, ``Deep
  spatial autoencoders for visuomotor learning,'' in \emph{2016 IEEE
  International Conference on Robotics and Automation (ICRA)}.\hskip 1em plus
  0.5em minus 0.4em\relax IEEE, 2016, pp. 512--519.

\bibitem{oh2017value}
J.~Oh, S.~Singh, and H.~Lee, ``Value prediction network,'' in \emph{Advances in
  Neural Information Processing Systems}, 2017, pp. 6118--6128.

\bibitem{silver2017predictron}
D.~Silver, H.~Hasselt, M.~Hessel, T.~Schaul, A.~Guez, T.~Harley,
  G.~Dulac-Arnold, D.~Reichert, N.~Rabinowitz, A.~Barreto, \emph{et~al.}, ``The
  predictron: End-to-end learning and planning,'' in \emph{International
  Conference on Machine Learning}.\hskip 1em plus 0.5em minus 0.4em\relax PMLR,
  2017, pp. 3191--3199.

\bibitem{muzero}
J.~Schrittwieser, I.~Antonoglou, T.~Hubert, K.~Simonyan, L.~Sifre, S.~Schmitt,
  A.~Guez, E.~Lockhart, D.~Hassabis, T.~Graepel, \emph{et~al.}, ``Mastering
  atari, go, chess and shogi by planning with a learned model,'' \emph{arXiv
  preprint arXiv:1911.08265}, 2019.

\bibitem{pathak2018zero}
D.~Pathak, P.~Mahmoudieh, G.~Luo, P.~Agrawal, D.~Chen, Y.~Shentu, E.~Shelhamer,
  J.~Malik, A.~A. Efros, and T.~Darrell, ``Zero-shot visual imitation,'' in
  \emph{Proceedings of the IEEE Conference on Computer Vision and Pattern
  Recognition Workshops}, 2018, pp. 2050--2053.

\bibitem{metaworld}
T.~Yu, D.~Quillen, Z.~He, R.~Julian, K.~Hausman, C.~Finn, and S.~Levine,
  ``Meta-world: A benchmark and evaluation for multi-task and meta
  reinforcement learning,'' in \emph{Conference on Robot Learning}, 2020, pp.
  1094--1100.

\bibitem{sermanet2016unsupervised}
P.~Sermanet, K.~Xu, and S.~Levine, ``Unsupervised perceptual rewards for
  imitation learning,'' \emph{arXiv preprint arXiv:1612.06699}, 2016.

\bibitem{sermanet2017time}
P.~Sermanet, C.~Lynch, J.~Hsu, and S.~Levine, ``Time-contrastive networks:
  Self-supervised learning from multi-view observation,'' in \emph{2017 IEEE
  Conference on Computer Vision and Pattern Recognition Workshops
  (CVPRW)}.\hskip 1em plus 0.5em minus 0.4em\relax IEEE, 2017, pp. 486--487.

\bibitem{nair2017combining}
A.~Nair, D.~Chen, P.~Agrawal, P.~Isola, P.~Abbeel, J.~Malik, and S.~Levine,
  ``Combining self-supervised learning and imitation for vision-based rope
  manipulation,'' in \emph{2017 IEEE International Conference on Robotics and
  Automation (ICRA)}.\hskip 1em plus 0.5em minus 0.4em\relax IEEE, 2017, pp.
  2146--2153.

\bibitem{zhou2019watch}
A.~Zhou, E.~Jang, D.~Kappler, A.~Herzog, M.~Khansari, P.~Wohlhart, Y.~Bai,
  M.~Kalakrishnan, S.~Levine, and C.~Finn, ``Watch, try, learn: Meta-learning
  from demonstrations and reward,'' \emph{arXiv preprint arXiv:1906.03352},
  2019.

\bibitem{taco}
K.~Shiarlis, M.~Wulfmeier, S.~Salter, S.~Whiteson, and I.~Posner, ``Taco:
  Learning task decomposition via temporal alignment for control,'' \emph{arXiv
  preprint arXiv:1803.01840}, 2018.

\bibitem{metarl}
Y.~Duan, X.~Chen, R.~Houthooft, J.~Schulman, and P.~Abbeel, ``Benchmarking deep
  reinforcement learning for continuous control,'' in \emph{International
  Conference on Machine Learning}, 2016, pp. 1329--1338.

\bibitem{rl2}
Y.~Duan, J.~Schulman, X.~Chen, P.~L. Bartlett, I.~Sutskever, and P.~Abbeel,
  ``Rl$^2$: Fast reinforcement learning via slow reinforcement learning,''
  2016.

\bibitem{irl}
A.~Y. Ng, S.~J. Russell, \emph{et~al.}, ``Algorithms for inverse reinforcement
  learning.'' in \emph{Icml}, vol.~1, 2000, p.~2.

\bibitem{irl2}
P.~Abbeel and A.~Y. Ng, ``Apprenticeship learning via inverse reinforcement
  learning,'' in \emph{Proceedings of the twenty-first international conference
  on Machine learning}, 2004, p.~1.

\bibitem{kelley1960gradient}
H.~J. Kelley, ``Gradient theory of optimal flight paths,'' \emph{Ars Journal},
  vol.~30, no.~10, pp. 947--954, 1960.

\bibitem{schmidhuber1990line}
J.~Schmidhuber, ``An on-line algorithm for dynamic reinforcement learning and
  planning in reactive environments,'' in \emph{1990 IJCNN international joint
  conference on neural networks}.\hskip 1em plus 0.5em minus 0.4em\relax IEEE,
  1990, pp. 253--258.

\bibitem{henaff2017model}
M.~Henaff, W.~F. Whitney, and Y.~LeCun, ``Model-based planning in discrete
  action spaces,'' \emph{arXiv preprint arXiv:1705.07177}, 2017.

\bibitem{attention}
A.~Vaswani, N.~Shazeer, N.~Parmar, J.~Uszkoreit, L.~Jones, A.~N. Gomez,
  L.~Kaiser, and I.~Polosukhin, ``Attention is all you need,'' \emph{arXiv
  preprint arXiv:1706.03762}, 2017.

\end{thebibliography}


\end{document}